\documentclass[letterpaper, 10 pt, conference]{ieeeconf}  

\IEEEoverridecommandlockouts         

\usepackage{graphics} 
\usepackage{graphicx}

\usepackage{booktabs}

\usepackage{siunitx}    
\usepackage{algorithm}
\usepackage{float}
\usepackage{amsmath} 
\usepackage{amssymb}

\usepackage{cite}
\usepackage{balance}
\title{Mapless Collision-Free Flight via MPC using Dual KD-Trees in Cluttered Environments}
\author{
Linzuo Zhang, Yu Hu, Yang Deng, Feng Yu, and Danping Zou$^{\dag}$%
\thanks{$^{\dag}$ Corresponding author. All authors are with the Shanghai Key Laboratory of Navigation and Location-Based Services, School of Sensing Science and Engineering, Shanghai Jiao Tong University. This work was supported in part by
the National Key R\&D Program of China under Grant 2022YFB3903802 and
in part by the Shanghai Municipal Collaborative Innovation Project under Grant HCXBCY-2023-029.}%
\thanks{\scriptsize \tt Emails: \{zhanglinzuo, henryhuyu, dengyang24, yu-feng, dpzou\}@sjtu.edu.cn}
}

\begin{document}
\maketitle
\thispagestyle{empty}
\pagestyle{empty}

\begin{abstract}
Collision-free flight in cluttered environments is a critical capability for autonomous quadrotors. Traditional methods often rely on detailed 3D map construction, trajectory generation, and tracking. However, this cascade pipeline can introduce accumulated errors and computational delays, limiting flight agility and safety. In this paper, we propose a novel method for enabling collision-free flight in cluttered environments without explicitly constructing 3D maps or generating and tracking collision-free trajectories. Instead, we leverage Model Predictive Control (MPC) to directly produce safe actions from sparse waypoints and point clouds from a depth camera. These sparse waypoints are dynamically adjusted online based on nearby obstacles detected from point clouds. To achieve this, we introduce a dual KD-Tree mechanism: the Obstacle KD-Tree quickly identifies the nearest obstacle for avoidance, while the Edge KD-Tree provides a robust initial guess for the MPC solver, preventing it from getting stuck in local minima during obstacle avoidance. We validate our approach through extensive simulations and real-world experiments. The results show that our approach significantly outperforms the mapping-based methods  and is also superior to imitation learning-based methods, demonstrating reliable obstacle avoidance at up to 12 m/s in simulations and 6 m/s in real-world tests. Our method provides a simple and robust alternative to existing methods. The code is publicly available at https://github.com/SJTU-ViSYS-team/avoid-mpc.

\end{abstract}
 \section{Introduction}
Flight in cluttered environments without collision remains a fundamental challenge for autonomous quadrotors. Traditional approaches typically employ a multi-stage hierarchical framework \cite{usenkoRealtimeTrajectoryReplanning2017, zhouEGOPlannerESDFFreeGradientBased2021, hanFIESTAFastIncremental2019}: First, environmental data are integrated through explicit map representations such as occupancy grids or Euclidean Signed Distance Fields (ESDFs)\cite{ding2019efficient, zhouRobustEfficientQuadrotor2019,zhou2021raptor}. A collision-free path is subsequently generated, which is then parameterized as trajectories (e.g., polynomials or B-splines) and tracked via low-level controllers ensuring smoothness and dynamic feasibility.

While effective in real-world cluttered environments, this cascade paradigm faces two inherent limitations. First, the accumulated error from each of the sequential modules including occupancy mapping,  path planning, trajectory generation, and trajectory tracking could produce biased actions making the quadrotor crash\cite{sunComparativeStudyNonlinear2022}. Second, the sequential workflow also introduces significant computational delays, impairing responsiveness in high-speed scenarios\cite{luYouOnlyPlan2024}.  Both limitations severely restrict the agility and safety of quadrotors. It naturally raises the question of whether approaches without mapping or trajectory generation exist, and whether agility and safety can be significantly enhanced.

    \begin{figure}[t]
        \centering
        \includegraphics[width=0.95\linewidth]{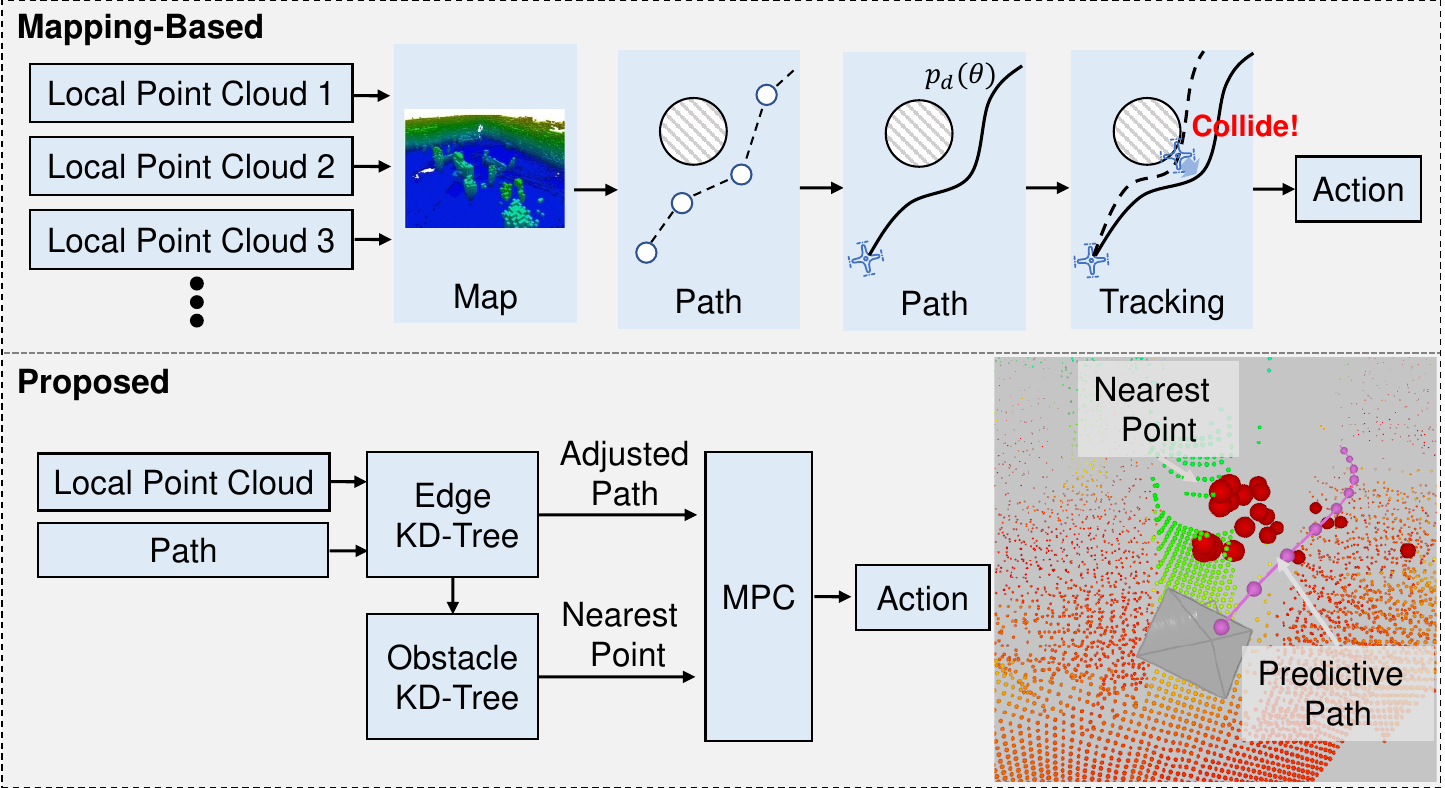}
        \caption{{\bf Comparison of the mapping-based approach and the proposed framework for obstacle avoidance:} The traditional mapping-based approach follows a pipeline consisting of mapping, path generation, trajectory optimization, and trajectory tracking. In contrast, the proposed framework optimizes obstacle avoidance through MPC while directly utilizing the path for efficient, real-time navigation.}
        \label{fig:fig1}
    \end{figure}
    \begin{figure*}[!thp]
        \centering
        \includegraphics[width=0.85\linewidth]{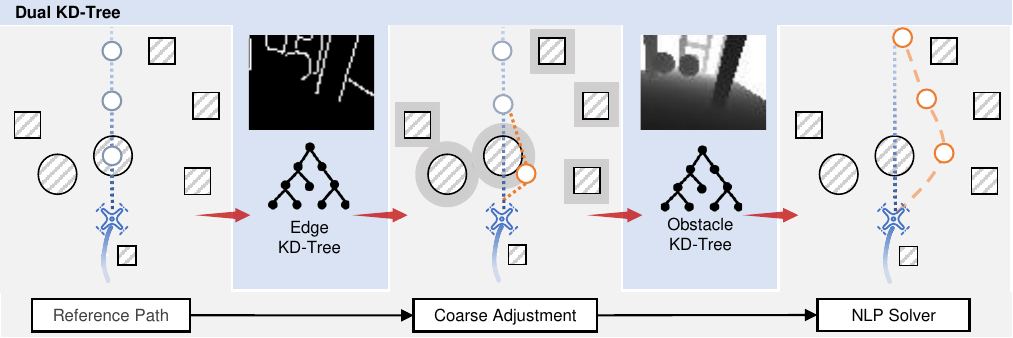}
        \caption{{\bf System Overview.} An MPC-based obstacle avoidance framework utilizes dilated edge maps to construct an edge KD-tree for coarse adjustment of the reference path and performs KNN-based nearest obstacle queries in the obstacle KD-tree to compute collision costs, which are optimized through nonlinear programming (NLP) for real-time control commands.}
        \label{fig:overview}
    \end{figure*}
 
    


To address these challenges, we propose a model predictive control (MPC) \cite{schwenzer2021review,romero2024actor, shin2021diffusing, nguyenTinyMPCModelPredictiveControl2024, salzmannRealtimeNeuralMPCDeep2023a,song2022policy} framework that integrates real-time local perception into the control loop. We use the local point cloud from the depth camera as input and solve for collision-free control actions using MPC, based on  sparsely sampled waypoints toward the goal. By integrating a novel mechanism to identify the nearest obstacle to each waypoint and incorporating this information into the MPC optimization, we ensure responsive and robust collision avoidance using only local perception data.

Specifically, we introduce a dual KD-tree \cite{tiwari2023developments,lu2023research,hamid2024nearest, florenceNanoMapFastUncertaintyAware2018} mechanism, as illustrated in Fig.~\ref{fig:fig1}. The Obstacle KD-Tree enables real-time nearest-neighbor queries for collision cost calculation, while the Edge KD-Tree optimizes waypoint adjustment by integrating dilated obstacle boundaries, providing robust initial guesses for the MPC solver to improve optimization efficiency and avoid local minima. This dual-KD tree structure not only simplifies the computational pipeline but also enhances the framework's ability to handle complex environments, ensuring both dynamic feasibility and real-time performance. Our method eliminates the need for explicit trajectory generation, tracking, or complex mapping processes, significantly enhancing the agility and robustness of autonomous flight in cluttered environments.

Our contributions are summarized as follows:
    \begin{enumerate}
        \item We propose a concise and straightforward MPC framework that directly uses the nearest point of discrete waypoints for obstacle avoidance.

        \item We propose a dual KD-tree perception module that enables real-time nearest-neighbor queries for collision cost computation, while providing a good initial guess for the MPC, enhancing its optimization efficiency and preventing local optima.

        \item We validate our approach through extensive simulations and real-world experiments, results demonstrate it outperforms both mapping-based and imitation-learning-based methods, achieving high-speed obstacle avoidance of up to 12 m/s in simulations and 6 m/s in real-world tests.
    \end{enumerate}
\section{Related Work}

\subsection{Mapless Obstacle Avoidance Methods}

Various approaches enable autonomous quadrotor flights without maps or predefined trajectories. A similar approach to ours is the reactive method, which relies solely on point clouds for local planning. Florence et al. \cite{florence2020integrated} use depth information to select optimal trajectories from predefined motion primitives. Kong et al. \cite{kong2021avoiding} maintain a local KD-tree and use kinodynamic A* to compute safe trajectories toward target positions. Although these methods eliminate map dependency and maintain low computational complexity, their reliance on tracking constrained motion primitives constrains maneuverability in complex environments. Ren et al. \cite{ren2022bubble} developed the Bubble Planner to generate real-time, high-speed, collision-free trajectories using a sampling-based corridor method with the Receding Horizon Corridor strategy. However, this approach remains fundamentally dependent on explicit trajectory generation pipelines, while introducing a persistent requirement for dedicated tracking controllers to ensure trajectory following.

Recent learning-based methods \cite{geles2024demonstrating, xing2024contrastive,zhaoLearningAgilityAdaptation2024, xing2024multi} circumvent the need for map construction in local obstacle avoidance by directly processing sensory inputs such as camera images. The Agile framework \cite{loquercioLearningHighspeedFlight2021} applies imitation learning to generate trajectories from depth images and states. MAVRL \cite{yu2024mavrl} maps depth images to acceleration commands using reinforcement learning. Zhang et al. \cite{zhang2024back,hu2024seeing} propose a differentiable pipeline that computes dynamics gradients to update policies, achieving notable performance by calculating collision loss from the nearest obstacles during training. While these methods improve efficiency, their implicit reasoning lacks interpretability, failing to validate trajectory safety explicitly (e.g., dynamic feasibility, obstacle distance), risking local minima traps or disturbances in deployment.

\subsection{Obstacle Avoidance with Model Predictive Control}
While MPC excels in high-maneuverability UAV control, its computational efficiency limits its ability to integrate environmental sensing data (e.g., lidar or cameras). One approach is to exclude environmental information from MPC, relying on predefined obstacles. Lindqvist et al. \cite{lindqvist2020nonlinear} integrated dynamic obstacle trajectories into optimization but only considered a single obstacle with a motion capture system, limiting real-world applicability. Similarly, Liu et al. \cite{liu2023tight} employed an MPC-based strategy to optimize collision probability but only considered one obstacle. Ji et al. \cite{ji2021cmpcc, xu2022dpmpc} require maps for collision detection or corridor generation. Krinner et al. \cite{krinner2024time} extended MPCC with flight zone constraints but relied on gate information, limiting safety to fixed race tracks. Some methods integrate visual information but require extra modules for predefined obstacle detection. U-map-based \cite{oleynikova2015reactive} detection was implemented by Lin et al. \cite{linRobustVisionbasedObstacle2020} and Xing et al. \cite{xingAutonomousPowerLine2023}, with the latter using ellipsoidal representations for power line tracking. 
Although \cite{liu2023tight, linRobustVisionbasedObstacle2020, xingAutonomousPowerLine2023} have demonstrated that it is possible to enforce geometric constraints in the control loop to avoid obstacles, they rely on oversimplified obstacle representations (e.g., spherical or cylindrical proxies) to make the computation tractable \cite{olcay2024dynamic, wu2024GPUACC}. Additionally, these methods depend on explicit obstacle detection, which is usually limited to predefined classes, such as pedestrians and pillars. Therefore, their approaches cannot be easily extended to cluttered environments, which are typically filled with unseen obstacles of irregular 3D shapes. 



\section{System Overview}
\label{subsec:preprocessing}
The proposed MPC-based autonomous flight system, as shown in Fig. \ref{fig:overview}, consists of two main modules: (1) a dual KD-tree module, where the Obstacle KD-Tree computes collision costs and the Edge KD-Tree adjusts waypoints to refine the initial optimization values, and (2) an MPC-based optimization module for real-time planning and control.

The system begins by sparsely sampling a fixed number of waypoints from the current position toward the goal. It then constructs the Obstacle KD-Tree and Edge KD-Tree using the depth map from the RGB-D camera and adjusts the waypoints if a potential collision is detected. The Obstacle KD-Tree stores obstacle points extracted from the depth map. While the Edge KD-Tree is built from a dilated, edge-filtered version of the depth map. Next, the MPC optimization module refines the waypoints generated by the Edge KD-Tree, incorporating kinematic constraints and collision costs. The optimization process minimizes a cost function over a fixed finite time horizon, producing an optimal sequence of control inputs and predictive waypoints. The resulting control commands are then sent to the quadrotor’s low-level flight controller (FCU). This entire process repeats at each time step.

\section{Dual KD-Tree Perception Module}

Below, we provide a detailed description of the construction and query process of the dual KD-Tree module. The input depth map is first downsampled to a resolution of \(64 \times 48\) pixels. This resolution was selected as a trade-off between runtime efficiency and sufficient obstacle detail. Two spatially indexed structures are then constructed to enable efficient collision checking and trajectory refinement:

\subsubsection{Edge KD-Tree and Waypoint Coarse Adjustment}  
\begin{figure}[h]
    \centering
    \includegraphics[width=\linewidth]{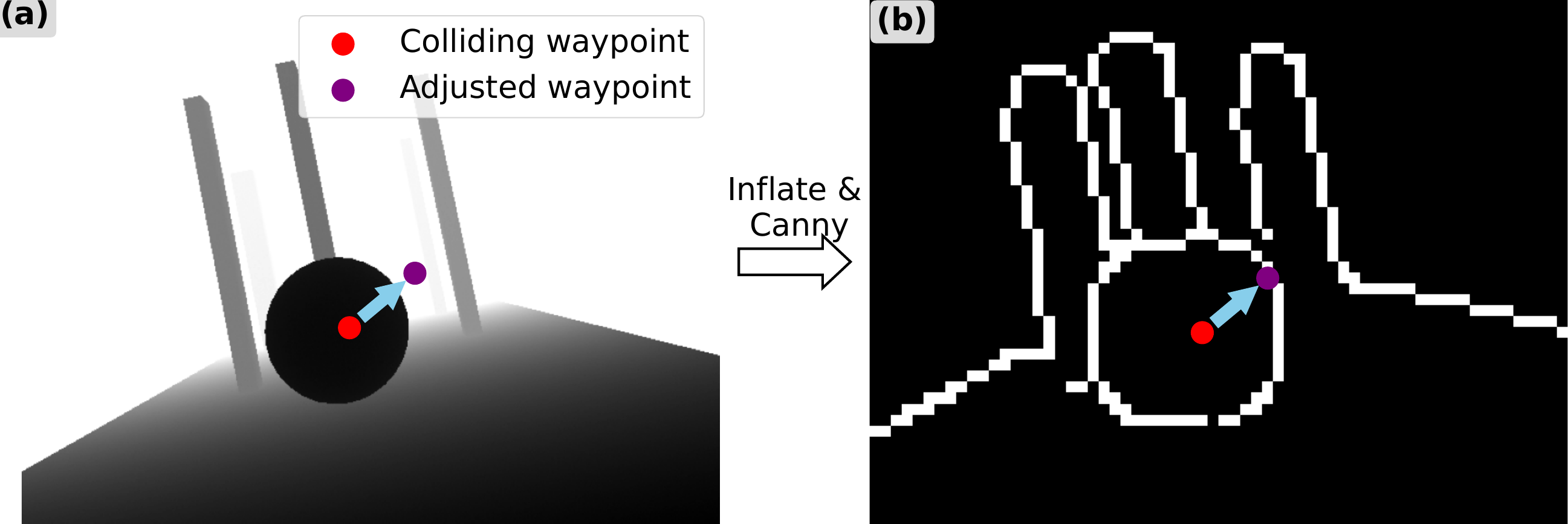}
    \caption{{\bf Waypoint Coarse Adjustment:}
    (a) Raw depth map;(b) Processed depth map after morphological inflation and Canny edge detection, enabling safe waypoint adjustment.}
    \label{fig:waypoint_adjust}
\end{figure}

The Edge KD-Tree is generated from a dilated and edge-filtered depth map, serving as a coarse adjustment layer for global waypoint adjustment. Dilation expands obstacle boundaries to: (1) provide safe-region initial waypoints for optimization, which effectively improves optimization efficiency; (2)  ensure that edge points mapped back to the Obstacle KD-Tree lie outside obstacles and near edges, thus preventing the nearest point from changing after optimization and avoiding local optima and (3) fill narrow gaps to prevent risky navigation. As shown in Fig. \ref{fig:waypoint_adjust}, during initial waypoint generation, colliding waypoints are replaced with their nearest edge points from the Edge KD-Tree to form a coarse collision-free path.

The inflation kernel size is determined by forward velocity and camera intrinsics. The inflation radius is computed as:  
\begin{equation}  
    r_{\text{infl}} = 2 \cdot \lfloor \frac{1}{2} \cdot \frac{d_s}{v_{\text{f}} \Delta t} \cdot f_{\text{pix}} \rfloor + 1  
\end{equation}  
where $\lfloor\cdot\rfloor$ is the floor function, \(d_s\) is the safe distance (derived from the quadrotor's radius and depth map noise), \(v_{\text{f}}\) is the forward velocity, \(\Delta t\) is the MPC optimization time step, and \(f_{\text{pix}}\) is the focal length in pixels.  
\subsubsection{Obstacle KD-Tree}
\label{subsubsec:obstacle_kd_tree}
This tree stores all obstacle points extracted from the downsampled depth map. During MPC optimization, the nearest obstacle point to each candidate waypoint is queried in real-time using nearest-neighbor search. The distance to this point is integrated into the collision cost function, ensuring obstacle avoidance during optimization.

Additionally, we employ a multi-frame strategy for enhancement. After constructing the current frame's KD-tree, we use the obstacle points from the previous keyframe to compare with the current frame in a separate thread. If the difference between the previous and current frame is too large, the current frame is designated as the new keyframe, and the previous frame is pruned by removing points that are too close to the current keyframe. Moreover, any keyframe entirely behind the current quadrotor position is also deleted. During queries, we prioritize searching for the nearest points within the current frame. If the waypoint is outside the current frame's field of view, the query is performed in parallel with the keyframes.


\section{Solving collision-free actions via MPC}

After using the edge KD tree to adjust the sparse waypoints according to the obstacle points, the MPC solver further takes these sparse waypoints by enforcing kinematic constraints and collision cost models and directly produces smooth, dynamically feasible control commands in real-time. 

\subsection{Optimization Formulation}
We formulate the collision-free flight problem as minimizing the cost function over a fixed time horizon $H$ through the following optimization framework:
\begin{align}
    \min_{u_{1:H}, x_{1:H}}\quad & 
    \mathcal{C}\left(x_{h}, u_{h}\right) 
    \\
    \text{s.t.}\quad             & \mathbf{x}_{h+1} = \text{RK4}\left(f_{d}\left(\mathbf{x}_{h}, \mathbf{u}_{h}\right), \Delta t\right) \notag                   \\
                                  & \mathbf{u}_{\text{min}} \leq \mathbf{u}_{h} \leq \mathbf{u}_{\text{max}} \notag                                                             \\
                                  & \mathbf{x}_{1} =\mathbf{x}_{\text{init}} \notag
\end{align}
The discrete-time state vector \(x_h = [\mathbf{p}_h, \phi_h, \mathbf{v}_h, \mathbf{a}_h]^\top\) is defined in the inertial frame, where: \(\mathbf{p}_h \in \mathbb{R}^3\) denotes the 3D position, \(\phi_h \in \mathbb{R}\) is the yaw angle, \(\mathbf{v}_h \in \mathbb{R}^3\) the linear velocity, and \(\mathbf{a}_h \in \mathbb{R}^3\) the linear acceleration. The control input \(\mathbf{u}_h = [\mathbf{a}_{c,h}, \phi_{c,h}]^\top\), defined in the inertial frame, comprises \(\mathbf{a}_{c,h} \in \mathbb{R}^3\) , and \(\phi_{c,h} \in \mathbb{R}\) (yaw angle command). The $\mathbf{u}_{\text{min}}$ and $\mathbf{u}_{\text{max}}$ define control input bounds, and $\mathbf{x}_{\text{init}}$ specifies the initial state.
The $\text{RK4}(\cdot)$ denotes the fourth-order Runge-Kutta method, while $f_d(\cdot)$ represents the quadrotor's dynamics model. After optimization, only the first action $\mathbf{u}_1$ is sent to the low-level flight control unit (FCU) to control the quadrotor. 


\subsection{Cost Function}

\label{subsec:cost_function}

The cost function  $\mathcal{C}\left(\mathbf{x}_{h}, \mathbf{u}_{h}\right)$ is a weighted sum of four terms: collision avoidance cost $\mathcal{C}_{a}$, terminal cost $\mathcal{C}_{goal}$, waypoint following cost $\mathcal{C}_{wp}$, and smoothness cost $\mathcal{C}_{u}$. The cost function is defined as:

\subsubsection{Collision Avoidance Cost}
The collision avoidance cost $\mathcal{C}_{a}(\mathbf{x}_{h})$ penalizes the proximity of the h-th waypoint to obstacles. We compute this cost using distances between the quadrotor position $\mathbf{p}_{h}$ and $M$ nearest obstacle points $\{\mathbf{o}_{h}^{i}\}_{i=1}^M$ queried from the Obstacle KD-Tree:
\begin{align}
    \mathcal{C}_{a}(\mathbf{x}_{h}) & = \lambda_c \sum_{i=1}^{M} \lVert \mathbf{v}_{h}^{i} \rVert\ln\left[1 + e^{\beta (r_q - \lVert \mathbf{p}_h - \mathbf{o}_h^i \rVert)}\right] \label{eq:avd_cost}
\end{align}
where the repulsion threshold $r_q\in\mathbb{R}_+$, obstacle-directed velocity $v_{h}^{i}\in\mathbb{R}^3$, and softplus-type barrier parameter $\beta\in\mathbb{R}_+$ constitute tunable parameters, with $\lambda_c\in\mathbb{R}_+$ scaling the cost's relative importance.

\subsubsection{Terminal Cost}
The terminal cost $\mathcal{C}_{goal}\left(\mathbf{x}_{H}, \mathbf{u}_{H}\right)$ is used to guide the quadrotor to the final goal point. This cost is computed with the goal state $\mathbf{x}_{\text{goal}}$:
\begin{equation}
    \mathcal{C}_{goal}\left(\mathbf{x}_{H}\right) = \left(\mathbf{x}_{H}- \mathbf{x}_{\text{goal}}\right)^{T}\mathbf{Q}_{\text{goal}}\left(\mathbf{x}_{H}- \mathbf{x}_{\text{goal}}\right)
\end{equation}
Here, $\mathbf{Q}_{\text{goal}}$ is a diagonal weighting matrix that adjusts the relative importance of the terminal cost. And the goal state $\mathbf{x}_{\text{goal}}$ is the final waypoint in the sequence.

\subsubsection{Discrete Waypoint Following Cost}
The discrete waypoint following cost, $\mathcal{C}_{wp}(\mathbf{x}_{h})$, encourages the discretized states $\mathbf{x}_{h}$ to closely follow the sparse waypoints $\mathbf{x}_{r}^{h}$:
\begin{equation}
    \mathcal{C}_{wp}(\mathbf{x}_{h}) = \left(\mathbf{x}_{h}- \mathbf{x}_{r}^{h}\right)^{T}\mathbf{Q}_{wp}\left(\mathbf{x}_{h}- \mathbf{x}_{r}^{h}
    \right)
\end{equation}
Here, $\mathbf{Q}_{wp}$ is a diagonal weighting matrix. Adjusting $\mathbf{Q}_{wp}$ influences how tightly the quadrotor follows the waypoints.

\subsubsection{Smoothness Cost}
The smoothness cost $\mathcal{C}_{u}(\mathbf{u}_{h})$ penalizes large control inputs $\mathbf{u}_{h}$, promoting smooth and energy-efficient control:
\begin{equation}
    \mathcal{C}_{u}(\mathbf{u}_{h}) = \mathbf{u}_{h}^{T}\mathbf{Q}_{u}\mathbf{u}_{h}
\end{equation}
where $\mathbf{Q}_{u}$ is a weighting matrix for control inputs.

\begin{figure}[h]
    \centering
    \includegraphics[width=0.9\linewidth]{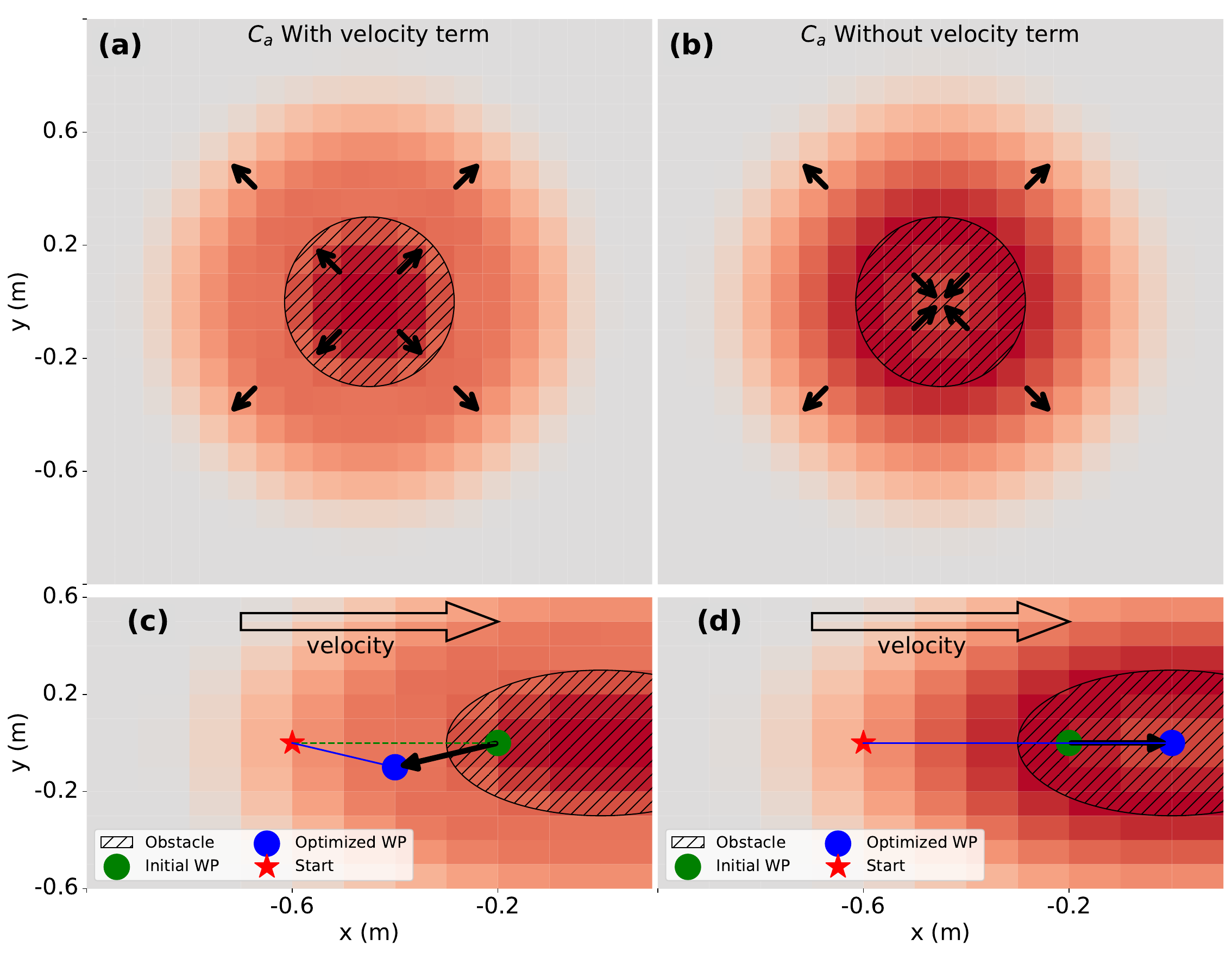}
    \caption{{\bf Distribution of Avoidance Costs:} (a) Cost function distribution with velocity term; 
    (b) Cost function distribution without velocity term; 
    (c) Path optimization process with velocity term; 
    (d) Path optimization process without velocity term. }
    \label{fig:costmap}
\end{figure}
\subsection{Analysis of the Collision Avoidance Cost}
The gradient of \eqref{eq:avd_cost} is derived as:
\begin{equation}
    \nabla \mathcal{C}_a = \sum_{i=1}^{M} \frac{-\beta \lVert \mathbf{v}_{h}^{i} \rVert e^{\beta (r_q - d_i)}}{1 + e^{\beta (r_q - d_i)}} \cdot \frac{\mathbf{p}_h - \mathbf{o}_h^i}{d_i}
    \label{eq:gradient}
\end{equation}
The gradient direction points from the obstacle to the waypoint(repulsive direction), and its relationship with the obstacle-directed velocity \( \mathbf{v}_h \in \mathbb{R}^3 \) requires specific analysis:
\begin{itemize}
    \item If the velocity direction is toward the obstacle (\( \mathbf{v}_h \cdot (\mathbf{p}_h - \mathbf{o}_h^i) < 0 \)), the gradient correction direction is opposite to the velocity direction, thereby mitigating collision;
    \item If the velocity is directed away from the obstacle, the gradient correction increases the velocity toward the obstacle. This scenario typically occurs when the waypoint appears inside the obstacle, as the nearest points are sampled on the obstacle surface and the obstacle is always located in front of the aircraft. In this case, the optimization process naturally pushes the waypoint out of the obstacle to ensure collision-free navigation.
\end{itemize}

Furthermore, for distances greater than $r_q$, the exponential term $e^{\beta (r_q - d_i)}$ rapidly decays toward zero due to the large choice of $\beta = 32$. Consequently, $\mathcal{C}_a$ is effectively negligible when the obstacle distance exceeds $r_q$,  This improves optimization efficiency, as the cost is large and decreases quickly within $r_q$, while after a few iterations, both the cost and the cost gradient become nearly zero.

Fig. \ref{fig:costmap} demonstrates how the velocity term in \eqref{eq:gradient} shapes avoidance cost fields. With velocity coupling (Fig. \ref{fig:costmap}a/c), $\mathcal{C}_a$ creates directional repulsion aligned with motion, pushing waypoints outward from obstacles (gray regions). Without it (Fig. \ref{fig:costmap}b/d), $\mathcal{C}_a$ forms radial symmetry and traps waypoints in central local minima. Additionally, as shown in the figure, the cost is primarily concentrated within a circular region of radius $r=0.6$ m around the obstacle, ensuring that the cost is negligible beyond this radius.indicating that our cost function is only effective within a specific range of the obstacle.
\subsection{Quadrotor Dynamics Model}
\begin{figure*}[!tp]
    \centering
    \includegraphics[width=0.95\linewidth]{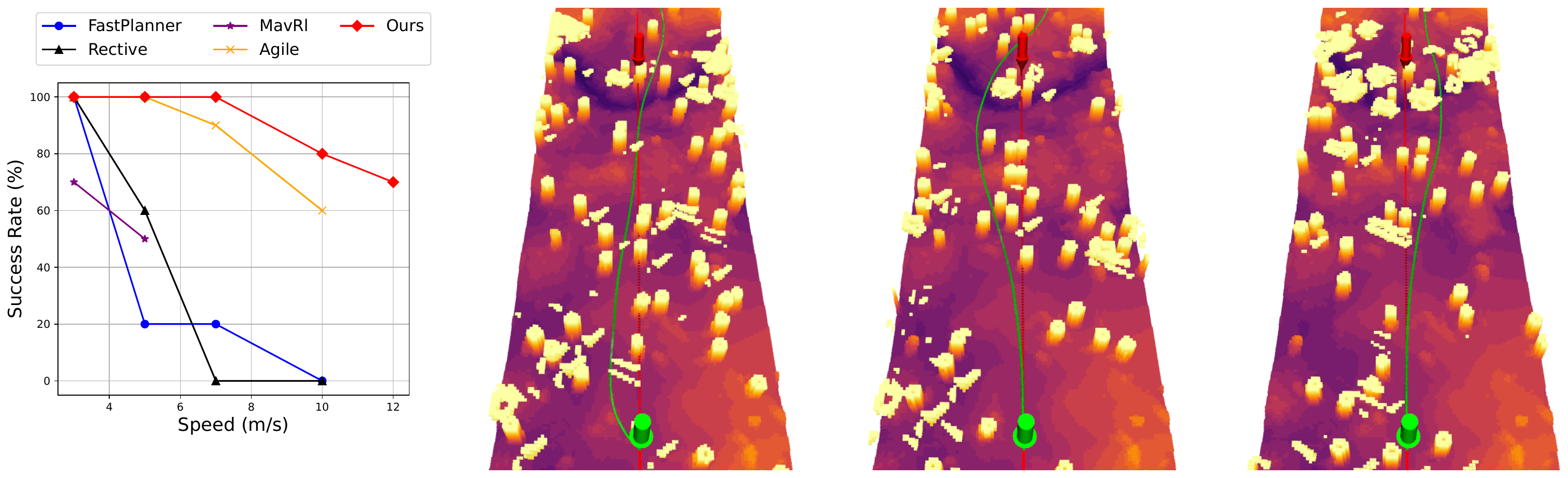}
    \caption{{\bf Performance Comparison of State-of-the-Art Methods in Flightmare Simulation:} On the left, we present the success rates of different methods at varying flight speeds, while the right figures illustrate the testing environments and the flight trajectories generated by our method.}
    \label{fig:sr_fm}
\end{figure*}
\begin{figure*}[!ht]
    \centering
    \includegraphics[width=0.95\linewidth]{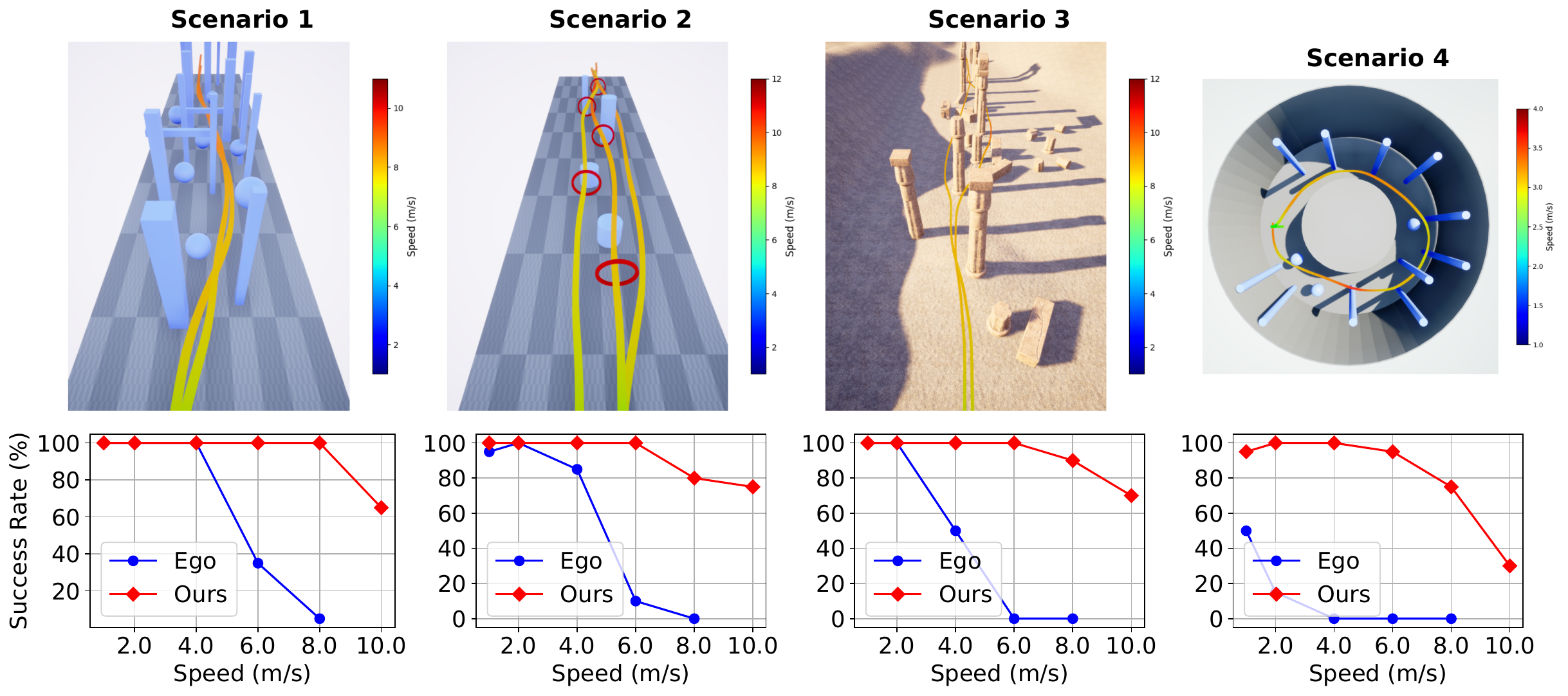}
    \caption{{\bf Performance of Our Method in Airsim Simulation:} On the top, we present the testing environments and the flight trajectories generated by our method. The bottom figures illustrate the success rates of our method and Ego-Planner\cite{zhouEGOPlannerESDFFreeGradientBased2021}  at varying desired speeds in different environments.}
    \label{fig:sr_airsim}
\end{figure*}
The state transition equation $\mathbf{x}_{h+1} = \text{RK4}(f_d(\cdot))$ is derived from the quadrotor's continuous-time dynamics model. In the following subsection, we provide a description of the quadrotor dynamics model used in $f_d(\cdot)$.

We model the quadrotor as a point mass and modeling the lower-level controller as a first-order system:
\begin{align}
    \mathbf{{\dot p}}_{h+1}   & =\mathbf{v}_{h} \notag                                                                  \\
    \mathbf{{\dot v}}_{h+1}    & =\mathbf{a}_h-\mathbf{R}_h^{T}\mathbf{D} \mathbf{R}_h \mathbf{v}_h \notag                                             \\
    \mathbf{{\dot a}}_{h+1}    & = \frac{\mathbf{K}_{a} \mathbf{a}_{c,h} - \mathbf{a}_h}{\mathbf{\tau}_{a}}\notag                        \\
    \mathbf{{\dot \phi}}_{h+1} & = \frac{k_{\phi}\phi_{c,h} - \phi_h}{\tau_{\phi}}\label{eq:dynamic}
\end{align}

Here, the matrix $\mathbf{R}_h \in \mathbb{R}^{3\times 3}$ denotes the rotation matrix, computed from the current orientation. The parameter $\mathbf{D} = \text{diag}(d_{x}, d_{y}, d_{z})$ represents linear damping, while $\mathbf{K}_{a}\in \mathbb{R}^{3\times 3}$, $\mathbf{\tau}_{a}\in \mathbb{R}^{3}$, $k_{\phi}\in \mathbb{R}$, and $\tau_{\phi}\in \mathbb{R}$ define the first-order dynamics, representing the system gains and time constants respectively.


\section{Experiment}

\subsection{Implementation Details} 
To validate the proposed method, we designed both simulation and real-world experiments to evaluate its performance and robustness in unknown and cluttered environments. Experimental parameters are: the optimizer is limited to a maximum of 10 iterations, and the number of nearest points selected, $M$, is set to 3. The MPC optimization horizon is set to 1 second, with a time step of 0.033 seconds. The repulsion threshold $r_q=1.0$ is set so that the UAV starts reacting to obstacles when they are within 1.0m, encouraging it to maintain a distance greater than 1.0m from obstacles. This value can be reduced in slower or denser environments. Safety distance $d_s = 0.15$ was determined according to the physical size of the UAV, which has a radius of approximately 0.1m. The NLP problem is solved by CasADi \cite{anderssonCasADiSoftwareFramework2019} using the IPOPT solver. Each iteration, thirty discrete waypoints are uniformly sampled between the current and goal positions.

\subsection{Simulation Experiment}
\subsubsection{Flightmare Simulation}
We benchmark our method against several state-of-the-art approaches using the Flightmare simulator\cite{song2021flightmare} in a high-density forest environment ($\delta$ = 1/25 tree $m^{-2}$). The methods compared include: Agile Autonomy \cite{loquercioLearningHighspeedFlight2021} (imitation learning and MPC for trajectory tracking), Reactive (sample-based mapless method), MAVRL \cite{yu2024mavrl} (learning-based, MPC for velocity tracking), and Fast-Planner \cite{zhouRobustEfficientQuadrotor2019} (mapping-based planning). Each method is tested at target speeds of 2, 5, 7, 10, and 12 m/s, with 10 trials per condition. A trial is considered successful if the quadrotor reaches the goal within a 5-meter radius without collision.

The reactive method \cite{florence2020integrated}, similar to ours, is mapless. It samples by evaluating collision probabilities and then uses MPC to follow the sampled trajectory. Its performance is acceptable at low speeds but significantly deteriorates beyond 5 m/s, with a notable drop in success rate. This limitation stems from the constrained set of motion primitives, which impedes high-maneuverability flight in complex environments.

Agile Autonomy \cite{loquercioLearningHighspeedFlight2021}, which employs imitation learning and MPC for collision-free trajectory tracking, our method demonstrates superior performance at higher speeds. At 8 m/s and 10 m/s, our method achieves success rates of 100\% and 70\%, respectively, while Agile Autonomy achieves 90\% and 60\%. This highlights the robustness of our tightly coupled perception-action pipeline in speed-critical scenarios.

We also compare against MAVRL \cite{yu2024mavrl}, a learning-based approach, which is theoretically limited in its maximum speed due to its training parameter settings and velocity strategy, and Fast-Planner \cite{zhouRobustEfficientQuadrotor2019}, a mapping-based planner, which exhibits significant performance degradation at higher speeds due to its reliance on the ESDF map construction and the precision of the underlying controller.
\begin{figure}[!thp]
    \centering
    \includegraphics[width=0.9\linewidth]{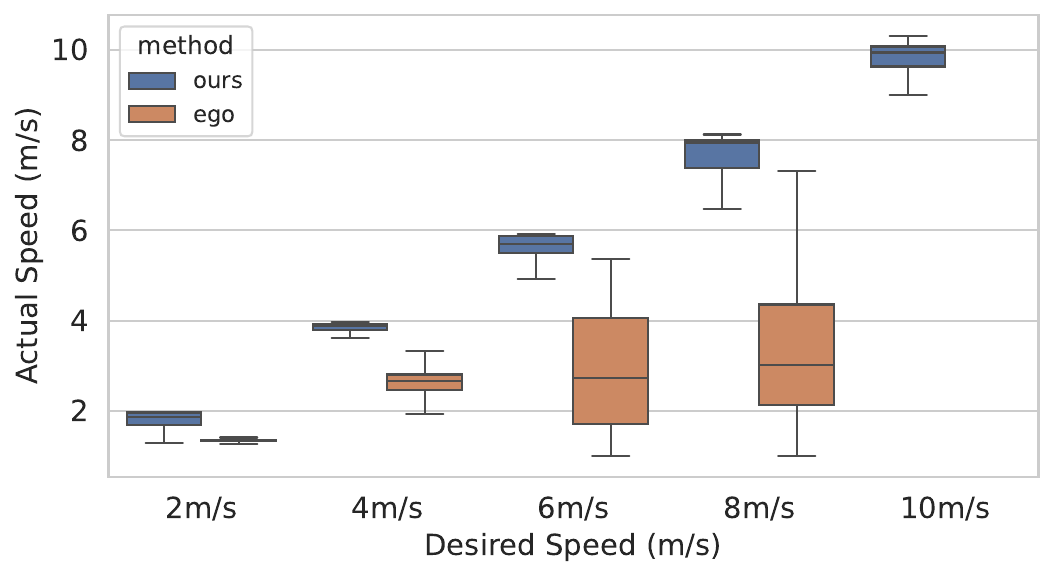}
    \caption{{\bf Speed Distribution of the Actual Forward Velocity:} We compare the actual forward velocity distribution between the proposed method and the Ego-Planner under different desired speeds in the AirSim simulation. The results show that our method maintains a more stable forward velocity across various desired speeds.}
    \label{fig:airsim_speed}
\end{figure}
\subsubsection{AirSim Simulation}
As shown in Fig. \ref{fig:sr_airsim}, we conducted experiments in AirSim-based environments across four scenarios: (1) A dense obstacle field with cubic, horizontal beam, and spherical obstacles to assess general and vertical avoidance; (2) A narrow, ring-shaped obstacle layout where the quadrotor navigated through ring centers at moderate speeds but bypassed laterally when rapid adjustments were needed; (3) A large-scale wilderness environment with cylindrical obstacles to test large obstacle avoidance; and (4) A circular obstacle-corridor scenario. In this scenario, forward arc length is calculated based on the known radius of the circular corridor and the desired velocity, which is then used to derive discrete reference waypoints. The quadrotor maintained tangential velocity alignment while continuously navigating the corridor with obstacle avoidance capability.

We compared our approach with the mapping-and-planning-based Ego-Planner\cite{zhouEGOPlannerESDFFreeGradientBased2021} across all scenarios. Quantitative results indicate that while Ego-Planner achieves satisfactory performance at very low speeds ($<3 m/s$), its success rate deteriorates markedly with increasing velocity targets - a trend consistent with findings from the original paper's real-world experiments. Our method maintains consistently high success rates across all tested scenarios and velocity regimes, demonstrating superior robustness and generalization capacity.

Our method can maintain a stable forward velocity, as shown in Fig. \ref{fig:airsim_speed}, we analyzed the distribution of actual forward velocity for different desired speeds in airsim simulation Scenario 1. The box plot illustrates that, compared to the Ego-Planner, our method consistently maintains a more stable forward velocity. While Ego-Planner demonstrates greater speed variance across all target velocities and struggles to attain desired speeds at higher ranges, our method achieves more consistent performance.

\begin{table}[!thp]
    \centering
    \caption{Ablation Study Results (Success Rate \%)}
    \label{tab:ablation}
    \begin{tabular}{l*{5}{c}}
    \toprule
    \textbf{Speed (m/s)} & \textbf{GT} & \textbf{NOISY}& \textbf{W/O Edge} & \textbf{Single} & \textbf{Single} \\
     &\textbf{Input} & \textbf{Input} & \textbf{KD-Tree} & \textbf{Frame} & \textbf{Nearest} \\
    \midrule
    3.0   & \textbf{100} & \textbf{100} & 80  & 90  & 90  \\
    5.0   & \textbf{100} & \textbf{100} & 70  & \textbf{100} & \textbf{100} \\
    7.0   & \textbf{100} & \textbf{100} & 60 & 90  & \textbf{100} \\
    10.0  & \textbf{90}  & 80 & 80  & 80  & \textbf{90} \\
    12.0  & \textbf{70}  & \textbf{70} & 60  & \textbf{70} & 60  \\
    \bottomrule
    \end{tabular}
\end{table}
\subsection{Ablation Experiment}
To evaluate the robustness of our method, we conducted an ablation study in the Flightmare simulator \cite{song2021flightmare}. As shown in Table~\ref{tab:ablation}, removing the Edge KD-Tree module significantly reduced the success rate, particularly at low speeds (e.g., 3 m/s), where the quadrotor passes through narrow gaps, and at high speeds (e.g., 12 m/s), where poor initial values hinder convergence to an optimal trajectory.

We also tested the system with a single frame, excluding the multi-frame strategy from Section~\ref{subsubsec:obstacle_kd_tree}. Results in the "Single Frame" columns of Table~\ref{tab:ablation} show a significant performance drop, emphasizing the importance of the multi-frame strategy for stability.

Additionally, using only one nearest point ($M=1$) in the Edge KD-Tree led to reduced performance, particularly with instability at 3 m/s (see "Single Nearest" column for 3.0 m/s). This suggests that multiple nearest points enhance robustness by providing more environmental data.
\begin{figure}[!thp]
    \centering
    \includegraphics[width=0.95\linewidth]{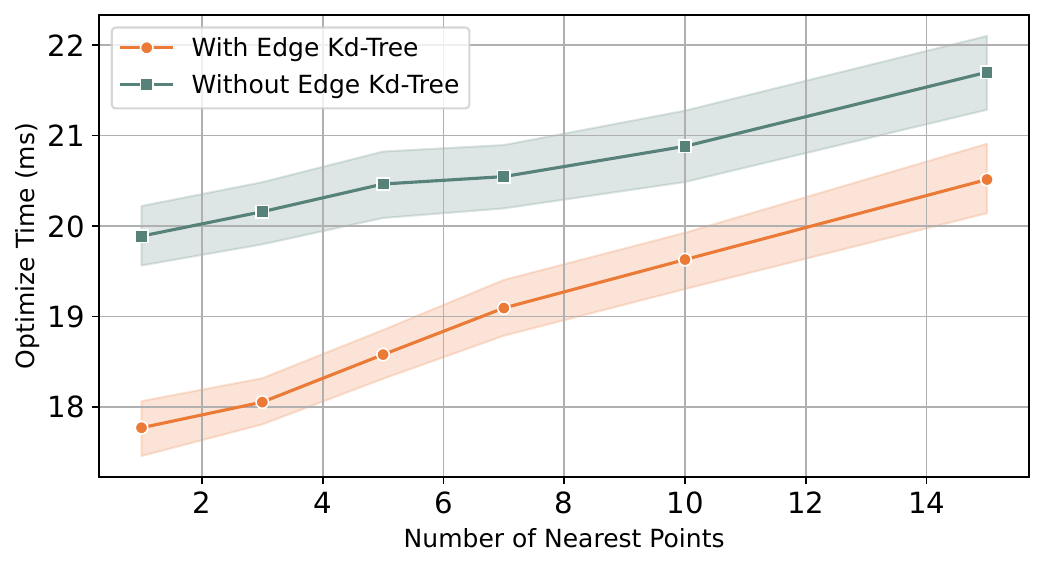}
    \caption{{\bf Curve of mean optimization time with varying nearest points:} A comparison of the time cost performance with and without the use of the Edge Kd-tree.}
    \label{fig:time-distribute-of-M}
\end{figure}
As shown in Fig.~\ref{fig:time-distribute-of-M}, the MPC optimization time increases linearly with the number of nearest points, impacting real-time performance. Therefore, we selected the three nearest points to balance performance and efficiency. Interestingly, the Edge KD-Tree reduced optimization time by providing a better initial guess.

We also conducted experiments to evaluate the impact of depth map and trajectory noise on the performance of our method. We repeated the same experiment in the Flightmare simulator \cite{song2021flightmare}, but provided all methods with depth maps computed from stereo pairs and trajectories that were perturbed with Gaussian noise. The depth maps were generated using the SGM algorithm, which inherently introduces realistic disparity noise, while zero-mean Gaussian noise with a standard deviation of 0.05m was added to trajectory positions to simulate localization errors. As shown in the first two columns of Table~\ref{tab:ablation}, our method is robust to depth map and trajectory noise. The results with noisy inputs are almost identical to those obtained without noise. This demonstrates the robustness of our method to sensor noise and inaccuracies in state estimation.

\subsection{Real-World Experiment}
We deployed our method on a lightweight real-world quadrotor platform (weights \SI{0.73}{\kilogram}) equipped with an Intel RealSense D435i depth camera and an Intel N100 onboard computer (4 cores, 3.4GHz). The platform demonstrates favorable thrust characteristics with a gravity-to-maximum-thrust ratio of 0.35, we integrated a millimeter-level precision motion capture system operating at 200 Hz while the low-level flight control was handled by a Betaflight controller.

\begin{figure}[!htp]
    \centering
    \includegraphics[width=\linewidth]{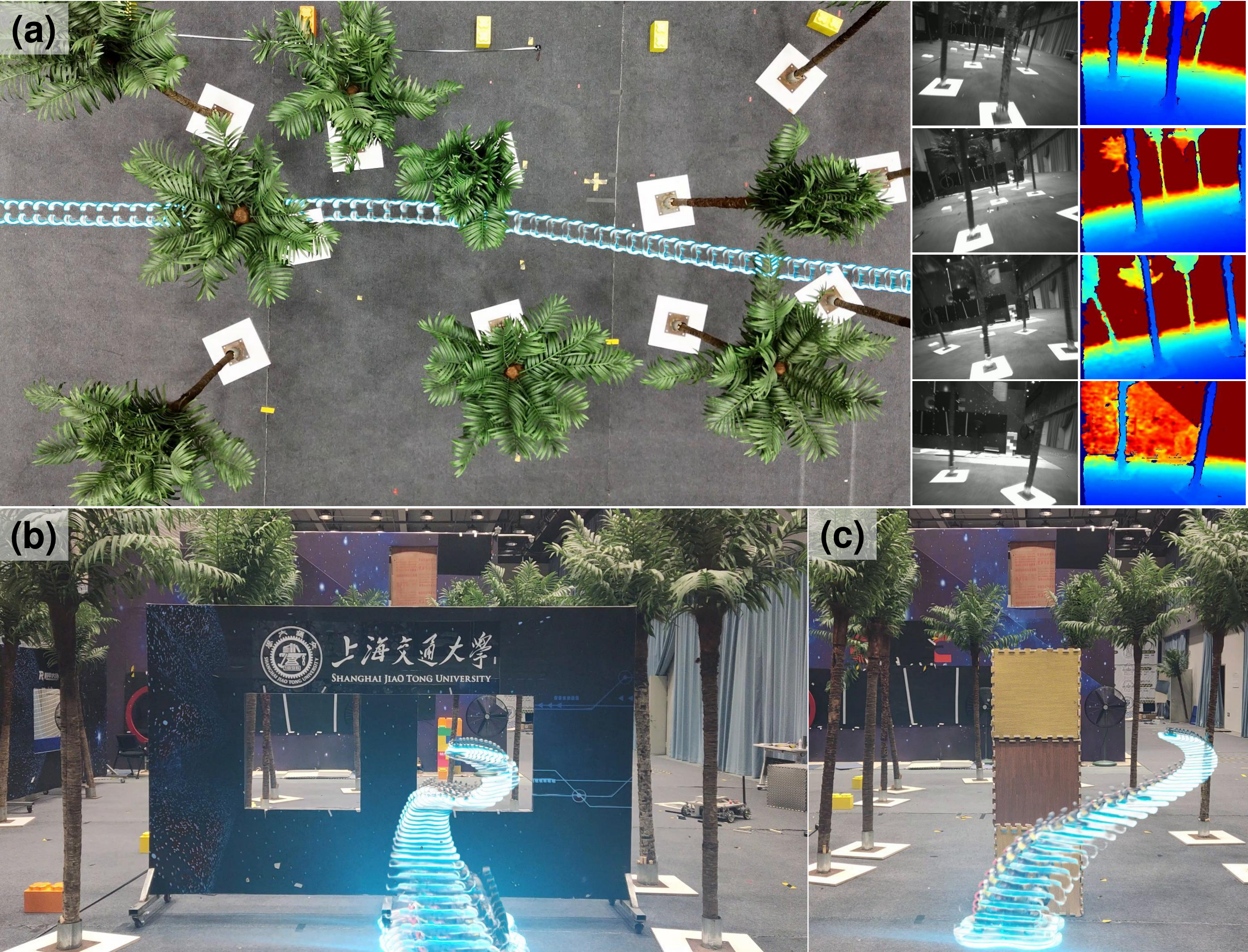}
    \caption{{\bf Real-world Experiment:} We validate our algorithm in various unknown environments. The quadrotor successfully navigates through these environments to reach the goal. The quadrotor's flight trajectory is visualized by overlaying its position in the video footage}
    \label{fig:real}
\end{figure}

We conducted experiments in various unknown environments, including a tree-dense areas with intricate branches and cluttered indoor settings, narrow-windowed door, and obstructed spaces with vertical pillars. The quadrotor successfully navigated through these environments to reach the goal, achieving a maximum speed of 6 m/s, as shown in \ref{fig:real}. These results demonstrate the robustness and generalization capabilities of our method in real-world scenarios.

We recorded the execution time of each module: the average time taken from inputting the depth map to constructing the KD-tree is 1.7ms, and the overall planning time averages 30ms. Among the planning tasks, the most time-consuming part is the NLP optimization solver, which has an average execution time of 18ms. The depth map input and planning tasks run synchronously on two separate threads. The output acceleration commands can reach a frequency of 30 Hz, which is sufficient to meet real-time control requirements.

\section{Conclusion}
\label{sec:conclusion}
The proposed method demonstrates the effectiveness of directly integrating point clouds into the MPC framework for obstacle avoidance. By leveraging the nearest obstacle points for each waypoint, this approach simplifies the computational pipeline and enhances scalability, thus enabling high-speed obstacle avoidance in complex environments. The proposed dual KD-tree module further streamlines obstacle detection and path adjustment, ensuring real-time performance and robustness. Simulations and real-world experiments validate the effectiveness of the approach, highlighting its potential for autonomous navigation in cluttered environments. 

This approach shares similarities in the core structure of the loss function (e.g., penalty terms based on exponential distance decay) with the end-to-end approach \cite{zhang2024back}, and relies on dynamically computing collision loss based on the nearest environmental points. However, the proposed method explicitly solves the solution through online numerical optimization without the need of training. Although the MPC-based approach still cannot compete this end-to-end approach in agility, it stands out as an important alternative, which offers better interpretability than the end-to-end approach. Future work will integrate learning-based strategies with optimization-based methods to balance agility, robustness, and interpretability.


\section*{Acknowledgement}

We thank SJTU SEIEE·G60 Yun Zhi AI Innovation and Application Research Center for experiment support. Special thanks are extended to Junjie Zhang for conducting real-world experimental validation, and to Qi Wu for valuable technical discussions that significantly enhanced this research.
\balance
\bibliographystyle{IEEEtran}
\bibliography{references}

\end{document}